\documentclass[letterpaper, 10 pt, conference]{ieeeconf}  

\IEEEoverridecommandlockouts                              

\renewcommand{\baselinestretch}{0.998}

\usepackage{amsmath}
\usepackage{amssymb}
\usepackage{tikz}

\usepackage{amsthm}
\usepackage{xspace}
\usepackage{multirow}
\usepackage{colortbl}
\usepackage{xcolor}
\usepackage{pgf}
\usepackage{ifthen}
\usepackage{calc}
\usepackage{setspace}
\usepackage{float}

\definecolor{lightblue}{RGB}{100,160,220} 
\definecolor{lightred}{RGB}{220,90,90}    

\newcommand{\colorcell}[1]{%
    \ifthenelse{#1 > 50}
        {\cellcolor{lightred!#1}{#1}}%
        {\cellcolor{lightblue!\number\numexpr 100-#1\relax}{#1}}%
}

\newcommand{\colorlow}[1]{%
    \ifthenelse{#1 > 5}
        {\cellcolor{lightred!#1}{#1}}%
        {\cellcolor{lightblue!\number\numexpr 100-#1\relax}{#1}}%
}

\newcommand{\midcell}[1]{%
    \begingroup
    \pgfmathsetmacro{\val}{#1}%
    \pgfmathtruncatemacro{\ival}{round(\val)}%
    \ifthenelse{\ival > 50}
        {\cellcolor{lightred!\ival}{\val}}%
        {\cellcolor{lightblue!\the\numexpr 100-\ival\relax}{\val}}%
    \endgroup
}

\usepackage{graphics} 

\usepackage[ruled,vlined]{algorithm2e} 
\usepackage[compact]{titlesec}
\titlespacing{\section}{0pt}{0.8ex plus .2ex minus .2ex}{0.3ex plus .1ex}

\newcommand{\madfull}{MADR: MPC-guided Adversarial DeepReach\xspace}
\newcommand{\mad}{MADR\xspace}
\usepackage[dvipsnames]{xcolor}
\usepackage{accents}
\usepackage{centernot}
\usepackage{cite}
\usepackage{subcaption} 
\usepackage[font=small,labelfont=bf]{caption}
\usepackage{diagbox} 
\newcommand{\state}{{x}}

\newcommand{\f}{f}
\newcommand{\fu}{g}
\newcommand{\fd}{w}

\newcommand{\Safe}{\mathcal{S}}
\newcommand{\Fail}{\mathcal{F}}
\newcommand{\traj}{\mathbf{x}}

\newcommand{\strategy}{\mathfrak{d}}
\newcommand{\Strategy}{\Xi}
\newcommand{\UU}{\mathcal{U}}

\newcommand{\ctrl}{u}
\newcommand{\dstb}{d}

\newcommand{\csig}{\mathbf{u}}
\newcommand{\dsig}{\mathbf{d}}

\newcommand{\tint}{\mathbb{T}}
\newcommand{\tinit}{t} 
\newcommand{\tvar}{s} 
\newcommand{\thor}{T} 
\newcommand{\cset}{\mathbb{U}}
\newcommand{\cfset}{\UU}
\newcommand{\dset}{\mathbb{D}}
\newcommand{\dfset}{\mathcal{D}}

\newcommand{\VV}{V}

\newif\ifmargincomments
\margincommentstrue
\ifmargincomments
\else

\fi

\newcommand{\stk}[1]{%
    \ifstrempty{#1}{}{
        \ifmmode
            \text{\sout{\ensuremath{#1}}}%
        \else
            \sout{#1}%
        \fi
    }%
}
\newcommand{\ICRAmode}{0}
\newcommand{\ICRAmod}[2]{%
  \ifcase\ICRAmode\relax
    #2\xspace
  \or
    \textcolor{red}{#2}\xspace
  \or
    \textcolor{red}{\stk{#1}#2}\xspace
  \fi
}

\newif\ifsuggestions
\suggestionstrue

\ifsuggestions
\newcommand\stnote[1]{\textcolor{cyan}{[ST: #1]}}
\newcommand\shnote[1]{\textcolor{violet}{[SH: #1]}}

\else
\newcommand\stnote[1]{}
\newcommand\shnote[1]{}
\fi

\usepackage{pgfplotstable}
\usepackage{booktabs} 
\usepackage{siunitx}  
\pgfplotsset{compat=1.18}

\pgfplotstableset{
    parse missing/.style={
        create on use/Parsed/.style={
            create col/expr={
                (\thisrow{#1}=="--")?nan:\thisrow{#1}
            }
        }
    }
}

\makeatletter
\let\NAT@parse\undefined
\makeatother
\usepackage{hyperref}

\makeatletter
\renewcommand{\@makecaption}[2]{%
  \vskip\abovecaptionskip
  \sbox\@tempboxa{\normalsize \textbf{#1}: #2}%
  \ifdim \wd\@tempboxa >\hsize
    \normalsize \setlength{\belowcaptionskip}{2pt} \textbf{#1}: #2\par
  \else
    \global \@minipagefalse
    \hb@xt@\hsize{\hfil\box\@tempboxa\hfil}%
  \fi
  \vskip\belowcaptionskip}
\makeatother

\title{\LARGE \bf
 \madfull
}

\author{Ryan Teoh$^{1,*}$, Sander Tonkens$^{2,*}$, William Sharpless$^{2}$, Aijia Yang$^{2}$, \\ Zeyuan Feng$^{3}$, Somil Bansal$^{3}$, and Sylvia Herbert$^{2}$ \\
\textcolor{MidnightBlue}{\href{https://land-dev.github.io/madr/}{https://land-dev.github.io/madr/}}
\thanks{$^1$University of California, Los Angeles, $^2$University of California, San Diego, $^3$Stanford University. $^{*}$Authors contributed equally. 
This work was supported by ONR under grants N00014-22-1-2292 and N00014-24-1-2661.
}
}

\begin{document}

\maketitle
\thispagestyle{empty}
\pagestyle{empty}

\begin{abstract}
Hamilton-Jacobi Reachability offers a framework for generating safe value functions and policies in the face of adversarial disturbance, but is limited by the curse of dimensionality. Physics-informed deep learning is able to overcome this infeasibility, but itself suffers from slow and inaccurate convergence, primarily due to weak PDE gradients and the complexity of self-supervised learning. Recent works have demonstrated that enriching the self-supervision process with regular supervision (based on the nature of the optimal control problem) greatly accelerates convergence and solution quality; however, these have been limited to single-player problems and simple games. 
In this work, we introduce \madfull, a general framework to robustly approximate the two-player, zero-sum differential game value function. In doing so, \mad yields the corresponding optimal strategies for both players in zero-sum games as well as safe policies for worst-case robustness. We test \mad on a multitude of high-dimensional simulated and real robotic agents with varying dynamics and games, finding that our approach significantly outperforms state-of-the-art baselines in simulation and produces impressive results in hardware. 
\end{abstract}

\section{Introduction}\label{sec:intro}
Zero-sum differential games provide a unifying framework for modeling interactions between an ego robot and an adversarial entity—whether that entity is a passive disturbance (e.g., wind, sensor noise) or an active agent (e.g., an opponent robot or human). 
To ensure robustness, zero-sum formulations assume that the disturbance or other player is adversarial, i.e., the ego robot aims to maximize a reward function (e.g., time to target, safety margin), whereas the disturbance or adversary attempts to minimize it.

Hamilton-Jacobi (HJ) Reachability is a popular approach for solving zero-sum differential games with respect to target achievement and obstacle avoidance \cite{mitchell2005HJ, chen2016fast}. HJ reachability requires solving a partial differential equation (HJ-PDE) to compute a value function; this value function implicitly provides a) the optimal control policy for the robot and adversary, and b) the reachable set: the set of states from which a system can reach (or avoid) a target (or obstacle) over time using these optimal policies. HJ reachability is particularly well-suited to safety-critical applications due to its ability to model worst-case scenarios in a zero-sum game-theoretic formulation~\cite{fisac2015reachavoid}.

\begin{figure}[h!]
    \centering
    \includegraphics[width=1.\columnwidth]{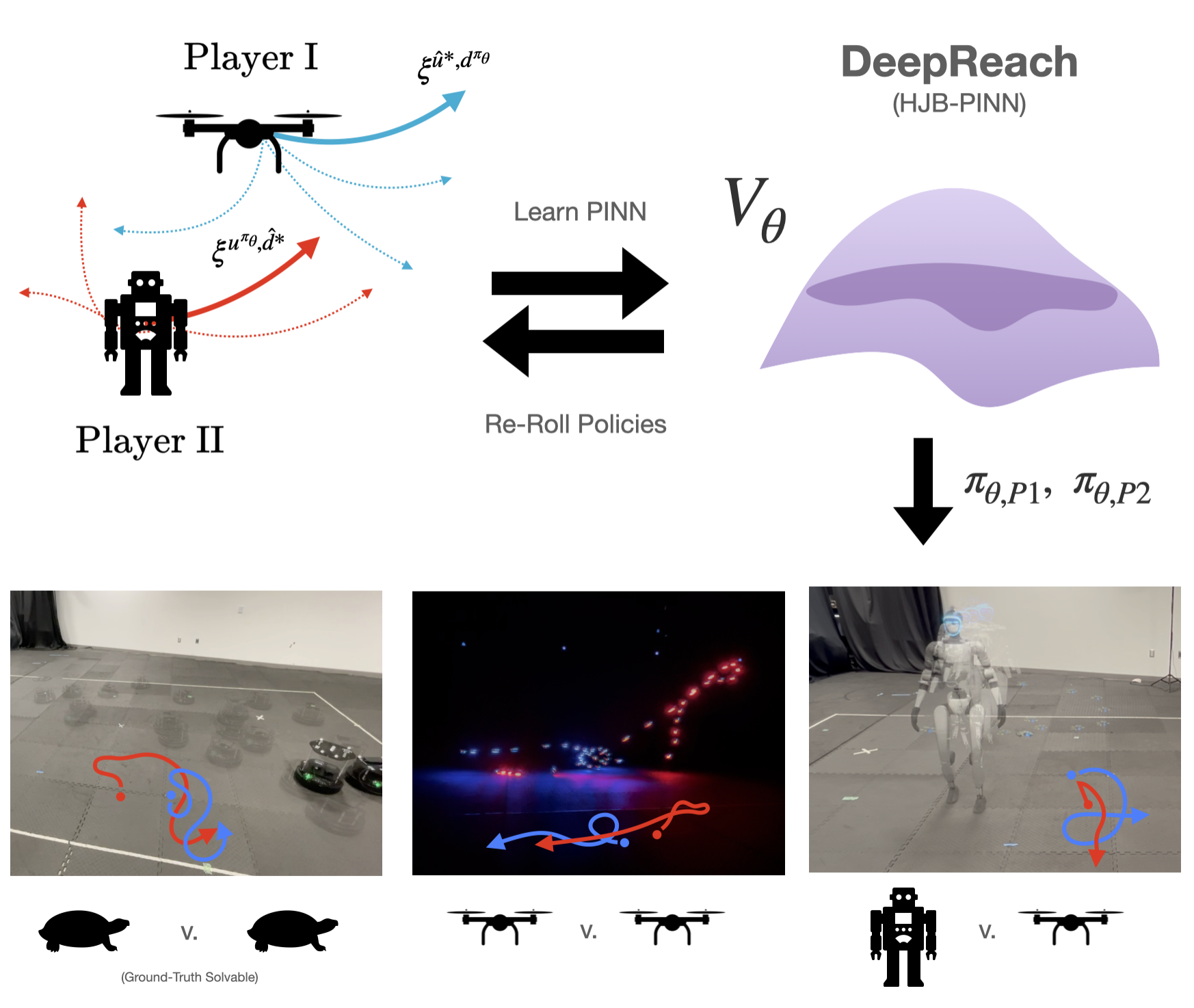}
    \label{fig:front_image.png}
    \vspace{-1em}

\caption{A graphical abstract of \mad and the robotic experiments. In this work, we propose enriching self-supervised learning of HJ-PDEs, i.e., DeepReach, with supervision given by the best sampled game rollout (top left), where the opponent's policy is defined by the current value approximation (top right). We demonstrate this approach performs across games with general dynamics (bottom), namely with TurtleBots, Drones, and Humanoid experiments.} \vspace{-1em}
\label{fig:frontpage}
\end{figure}

HJ reachability has been applied to a broad range of robotic systems, including collision avoidance for aerial vehicles, multi-agent coordination, and motion planning under uncertainty \cite{mitchell2005HJ,herbert2017fastrack}. However, classical approaches rely on grid-based dynamic programming and scale poorly with state dimension due to the curse of dimensionality~\cite{BansalChenEtAl2017b}. This limits their use to systems with fewer than six states in practice.

To address this challenge, learning-based approaches have been explored for scalability. Methods such as DeepReach \cite{bansal2021deepreach} replace the grid with a function approximator trained to satisfy the HJ-PDE in a physics-informed learning (PINN) framework, allowing scalability to more complex systems. 

DeepReach has primarily focused on control-only cases at high dimensions, with only limited work addressing robustness or more complex dynamics~\cite{Borquez_2023,TonkensShinde2025EtAl}. 
Recent works have demonstrated that introducing supervision greatly improves the learned approximation. In~\cite{sharpless2024linear}, the Hopf formula is used to solve a linearization of the game as a supervision loss, scaling to systems of 50 dimensions. However, this method struggles with nonlinear or angular dimensions, which frequently appear in robotics. 
More recently,~\cite{feng2025bridgingmodelpredictivecontrol} employed sampling-based Model Predictive Control (MPC) as a supervisory signal, finding that this significantly improves performance in high-dimensions but is limited to the control-only case. 

\subsection{Contributions} Our method, \mad, proposes solving zero-sum differential games using a value-only DeepReach approach combined with adversarial MPC roll-outs. 
Our key insight is to define the opponent's policy through the current value gradient approximation, which allows robust sampling of improved ego policies and the associated optimal cost of their trajectories.  
Unlike actor-critic methods, where policy performance depends on simultaneous learning of both actor and critic, our value-only approach ensures that high-quality policies emerge directly from accurately-learned value functions. 
To achieve this, we mitigate co-learning issues by collecting separate datasets in which the sampling (and associated adversarial gameplay under the current value gradient) occurs for each agent individually. 
This enables robust learning of a single value function for both players, leading to effective policies in adversarial scenarios, see Fig~\ref{fig:frontpage} for an overview. 

We empirically demonstrate that this value-informed adversarial supervision improves both the fidelity and safety margins of learned reachable sets, leading to better robustness against disturbances and adversarial agents. The effectiveness of \mad is validated through extensive simulation experiments and a variety of hardware demonstrations, highlighting improved safety performance in complex, high-dimensional systems exposed to significant disturbances or adversarial players. 
In addition, we compare \mad against other current approaches in head-to-head policy matchup comparisons.  Ultimately, our approach bridges the gap between the rigorous, theory-driven framework of HJ reachability and the practical use of optimal control to handle adversarial or worst-case scenarios in real-world environments. 

\subsection{Related Work} 

\textbf{Linear Quadratic Games (LQG)} are a class of differential games where multiple players aim to minimize a quadratic cost function over time, subject to linear dynamics, either cooperatively or non-cooperatively. In the non-cooperative scenario, players often seek Nash equilibria, which can be computed by solving coupled Riccati equations \cite{papavassilopoulos1979nash} in limited scenarios, but in practice, iteratively using sequential linearizations \cite{fridovich2020efficient}. The cooperative case typically involves joint optimization to minimize a shared cost function \cite{Zhao_2023}. Recent advancements have extended LQGs to incorporate uncertainties and stochastic disturbances for real-world scenarios \cite{falconi2025distributionally}. However, they remain relatively challenging in zero-sum settings and face limitations when applied to complex dynamical systems, where linearizing the dynamics leads to significant inaccuracies.

\textbf{Model Predictive Control (MPC)} is a model-based optimal control strategy. At each timestep, it solves a constrained optimization problem over a finite prediction horizon, a method known as receding horizon control. 
MPC is well-suited for safety-critical applications because it produces a control sequence that is guaranteed by the model to satisfy state and input constraints over the entire prediction horizon~\cite{BorrelliBemporadEtAl2017}. However, MPC has two significant drawbacks: its high computational burden and its sensitivity to model mismatch. 
Sampling-based variants, such as MPPI~\cite{williams2016mppi}, improve robustness to model inaccuracies by evaluating thousands of stochastic rollouts. 
Additionally, robust, stochastic, and distributed formulations further enhance performance and efficiency in uncertain environments~\cite{robustMPC}. 
Its high computational cost~\cite{schwenzer2021review} limits real-time usage during inference for safety-critical applications. 

\textbf{Adversarial Reinforcement Learning (RL)} formulates control in uncertain environments as a two-player game, where a controller learns to counteract worst-case disturbances or adversarial agents~\cite{pinto2017robust, li2023robust}. This framework improves robustness to unmodeled dynamics, environmental perturbations, and intelligent opponents. Modern algorithms, often based on actor-critic methods, enable scalable training in high-dimensional systems. However, these approaches are often brittle, as adversarial training can induce instability, high variance, or convergence issues~\cite{lowe2017multi}. 
Additionally, the reliance on additive payoffs in these methods makes them ill-suited for safety-critical tasks, thus motivating the integration of reachability for learning in such domains~\cite{FisacLugovoyEtAl2019, so2024solving}.

\textbf{ISAACS} (Iterative Soft Adversarial Actor-Critic for Safety) is a Safety Bellman-based representative framework that formulates safety-critical control as a zero-sum game between a control policy and a disturbance policy~\cite{hsu2023isaacs}. 
The disturbance acts as an adversary attempting to drive the system into unsafe regions, while the controller learns to maintain safety. 
ISAACS trains both players jointly via soft actor-critic methods. 
Extensions leverage the learned value function to implement a runtime safety filter~\cite{nguyen2024gameplayfiltersrobustzeroshot, wang2025magicsadversarialrlminimax}, overriding unsafe task policies with robust fallbacks generated from a learned safety policy.  
This framework enables real-time safety assurance in high-dimensional systems. 
However, ISAACS's reliance on actor-critic methods can make learning unstable or sample-inefficient in complex domains. 
Additionally, it prioritizes robustness against adversarial disturbances by optimizing the disturbance policy given a control policy, prioritizing conservativeness for the ego-agent compared to a zero-sum game formulation. 

\textbf{Organization:}
We begin in Section~\ref{sec:background} with a brief overview of HJ reachability analysis computation and online inference. 
In Section~\ref{sec:framework}, we introduce the proposed \madfull framework and detail the algorithmic approach for learning reachability solutions under adversarial MPC supervision. 
Section~\ref{sec:simulation} presents simulation results demonstrating the effectiveness of \mad, followed by hardware deployment results in Section~\ref{sec:hardware}. 
Finally, we summarize our findings in Section~\ref{sec:conclusion}.

\section{Preliminaries}\label{sec:background}

\subsection{Problem Formulation}

We consider a dynamical system with state $ \state \in \mathbb{R}^n$, control $ \ctrl \in \cfset $, and disturbance (or adversary) $ \dstb \in \dfset $, with $\cfset$ and $\dfset$ compact sets representing input constraints, governed by control-and-disturbance affine dynamics
\begin{equation}
\dot{\traj}(\tinit) = \f(\traj(\tinit)) + \fu(\traj(\tinit))\csig(\tinit) + \fd(\traj(\tinit))\dsig(\tinit),
\label{eq:dynamics}
\end{equation}
where $\traj$ is the state trajectory. $\csig:\tint \to \cfset$ and $\dsig:\tint \to \dfset$ are control and disturbance signals of $\cset$ and $\dset$ respectively, with $\tint=[\tinit,\thor]$ a finite time horizon. 
Let Player I choose $\ctrl$ while Player II chooses $\dstb$.

In this setting, Player I seeks to avoid entering a \emph{failure set} $ \Fail \subset \mathbb{R}^n $, despite the opposing actions of Player II. This failure set is the set of states that Player I must avoid to remain safe. This can include situations such as being captured by Player II or colliding with obstacles. Conversely, Player II acts adversarially to drive the system toward $ \Fail $. This interaction defines an \emph{avoid game}, where the objective of Player I is to maintain safety over the time horizon $\tint$. 

Throughout this work, we solve such games by characterizing the \emph{safe set} $ \Safe \subset \mathbb{R}^n $: the set of states from which Player I can guarantee safety against the worst-case disturbance within the specified horizon. In addition, we seek to obtain the optimal policies for both players. 

\subsection{Hamilton–Jacobi Formulation}

To characterize the safe set and corresponding optimal policies, we begin by defining a \textit{boundary function} $\ell(\state)$ that defines the failure set $\Fail \subset \mathbb{R}^n = \{\state:\ell(\state) \leq 0\}$. This is typically constructed as a signed distance function to $\Fail$. The objective function \ICRAmod{}{for our two player game} is defined as:
\begin{equation}
\label{eq:objective}
    J(\state,\tinit, \csig, \dsig) = 
\min_{\tvar\in[\tinit, \thor]} \, \ell \big(\traj_{x,t}^{\csig,\dsig}(\tvar)\big),
\end{equation} 
where $\traj_{x,t}^{\csig,\dsig}(\cdot)$ denotes the system trajectory starting from $\state$ at time $\tinit$ and state $\state$. 
The \textit{value function} $V(\state,t)$ encodes the safety of each state under worst-case disturbances and optimal control by optimizing over \eqref{eq:objective}. It is defined as
\begin{equation}
\label{eq:valuefunc}
    \VV(\state,\tinit) = 
    \min_{\strategy[\csig] \in \Strategy}
    \max_{\csig \in \cset} J(\state,\tinit, \csig, \strategy),
\end{equation}
where ${\strategy[\csig]}:\cset \to \dset$ denotes a nonanticipative strategy such that Player II may choose to play $\dsig$ based on the current and past values of the control $\csig$, but not on its future values. 

A positive value $\VV(\state,\tinit) > 0$ indicates that the robot starting from state $\state$ at time $\tinit$ can avoid entering the failure set under all admissible disturbances. Therefore, the \textit{safe set} at time $\tinit$ is given by $\Safe(\tinit) = \{ \state \in \mathbb{R}^n \,|\, \VV(\state,\tinit) > 0 \}.$

The value function is computed by solving a Hamilton–Jacobi-Isaacs variational inequality (HJI-VI):
\begin{equation}\label{eq:HJIVI}
\begin{split}
\min \Bigl\{ 
    &\frac{\partial \VV}{\partial \tinit}(\state,\tinit) 
    + \max_{\ctrl \in \cfset} \min_{\dstb \in \dfset} 
      \nabla \VV(\state,\tinit) \cdot \f(\state), \\
    &\ell(\state) - \VV(\state,\tinit)
\Bigr\} = 0, \hspace{4em} \VV(\state,\thor) = \ell(\state).
\end{split}
\end{equation}

The variational inequality ensures that once a state enters the failure set, it remains marked as unsafe for all earlier times, thus correctly capturing the avoid for all time game objective. 
Grid-based methods can compute the exact value function, but are intractable in high dimensions. 
We instead use a semi-supervised learning approach (see Sec.~\ref{sec:framework}) to approximate the value function, preserving key properties of HJ reachability while remaining computationally feasible.

\subsection{Online Control Using the Value Function}

Once the value function is computed offline, its gradients can be used online by each player to determine optimal control actions. Specifically, Player I chooses
$ \ctrl^*(\state,\tinit) = \arg\max_{\ctrl \in \mathcal{\cfset}} 
      \nabla_\state \VV(\state,\tinit) \cdot \fu(\state)\ctrl.$
Player II chooses
$\dstb^*(x,\tinit) = \arg\min_{\dstb \in \mathcal{\dfset}} 
      \nabla_\state \VV(\state,\tinit) \cdot \fd(\state)\dstb,
$
with $\fu(\state)$ and $\fd(\state)$ from~\eqref{eq:dynamics}.
The sign of the gradient $\nabla_\state \VV$ along these directions determines whether increasing or decreasing the input increases safety. 
Evaluating the gradient of a precomputed value function enables high-frequency real-time control, thus avoiding solving an optimization problem or rolling out many samples online. 
Optionally, this value function can act directly as a \emph{safety filter}, constraining control inputs to prevent entering unsafe regions under disturbances, see, e.g.,~\cite{BrunkeGreefEtAl2021} for details.

\section{Adversarial MPC-Guided Reachability Learning Framework}\label{sec:framework}

\begin{figure}[]
    \centering
    \includegraphics[width=1.0\columnwidth]{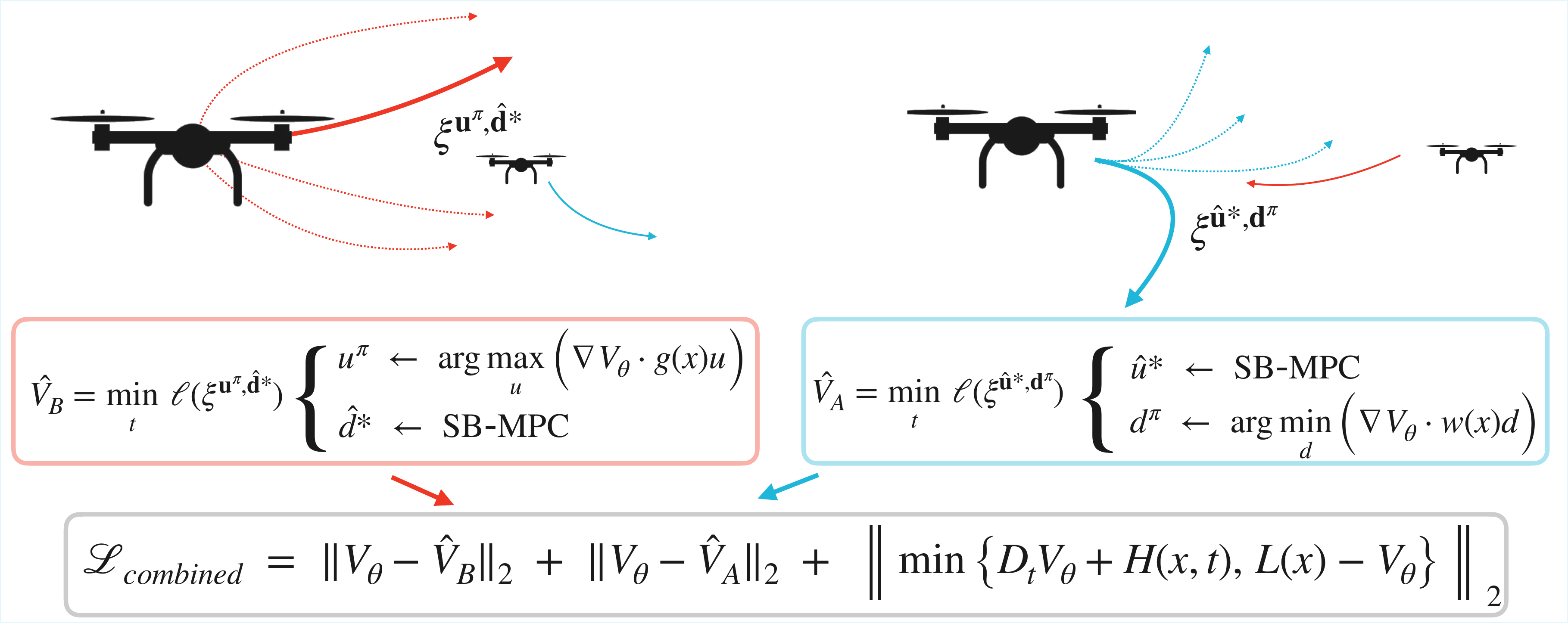} 
    \caption{Graphical depiction of the MPC formulation combining both players' rollouts in the loss function for the proposed MPG-guided adversarial PINN training.}
    \label{fig:mpc_formulation}
    \vspace{-1em}
\end{figure}

Building on the HJ reachability concepts introduced above, we present a framework for learning the 2-player value function in high-dimensional systems. 
Rather than solving~\eqref{eq:HJIVI} over a grid, our approach leverages self-supervised learning of the HJI-VI loss (using DeepReach~\cite{bansal2021deepreach}), enhanced with a supervised adversarial sampling-based MPC procedure that is generated in parallel to efficiently approximate the value function. 
We will first introduce the sampling-based MPC method that optimizes the objective~\eqref{eq:objective}. 
Then we describe how this dataset will be used to augment the training loss of DeepReach. 
These losses are combined, as illustrated in Fig~\ref{fig:mpc_formulation}, to guide the learned value function and its policies toward robust, safety-informed behavior in 2-player settings.

\begin{algorithm}[!t]
\setstretch{0.85}
\caption{Sampling-based MPC dataset [\textcolor{blue}{Control Perspective}] [\textcolor{red}{Disturbance Perspective}]}
\label{alg:robust_mpc_no_warmstart}

\KwIn{MPC dataset size $|D_\text{MPC}|$, horizon $H$, step size $\Delta t$, 
dynamics model $f$, learned value function $V_\theta$, constraint function $\ell$, 
sample size $N$, refinement iterations $K$}
\KwOut{$D_\text{MPC} = \{(\state_j,t_j,\hat{\VV}(\state_j, t_j))\}$}

\DontPrintSemicolon

\textbf{Initialization:} Set best cost \textcolor{blue}{$J^* \gets -\infty$} \textcolor{red}{$J^* \gets \infty$} $D_\text{MPC} = \emptyset$\;

\For{$j \gets 1$ \textbf{to} $|D_\text{MPC}|$}{
    $\state_j \sim \text{Uniform}(\mathcal{X}), \quad t_j \sim \text{Uniform}(0, \thor)$\;

    \For{$k \gets 1$ \textbf{to} $K$}{
        \For{$i \gets 1$ \textbf{to} $N$}{
            $\state_0^{(i)} \gets \state_j$

            \For{$h \gets 0$ \textbf{to} $H-1$}{
                \textcolor{blue}{$u^{(i)} \sim \mathcal{N}(\mu_\ctrl, \sigma_\ctrl^2)$}\;
                \textcolor{blue}{$d^{(i)} \gets \arg\min_\dstb \nabla V_\theta(x^{(i)}_h)\cdot w(x^{(i)}_h)\dstb$}\;

                \textcolor{red}{$\ctrl^{(i)} \gets \arg\max_\ctrl \nabla V_\theta(x^{(i)}_h)\cdot g(x^{(i)}_h)\ctrl$}\;
                \textcolor{red}{$\dstb^{(i)} \sim \mathcal{N}(\mu_\dstb, \sigma_\dstb^2)$}\;

                $x_{h+1}^{(i)} = x_h^{(i)} + \bar{\f}(x_h^{(i)})\Delta + \bar{\fu}(x_h^{(i)},\ctrl^{(i)})\Delta + \bar{\fd}(x_h^{(i)},\dstb^{(i)})\Delta$\;
            }

            $J^{(i)} = \min_h \ell(x_h^{(i)})$
            \If{$H < \tint$}{
                $V_\text{togo}^{(i)} \gets V_\theta(x_H^{(i)})$\;
                $J^{(i)} \gets \min\{J^{(i)}, V_\text{togo}^{(i)}\}$\;
            }

            \textbf{Update best cost:}\;
            \If{\textcolor{blue}{$J^{(i)} > J^*$} \textcolor{red}{$J^{(i)} < J^*$}}{
                \textcolor{blue}{$\mu_\ctrl \gets \ctrl^{(i)}$}
                \textcolor{red}{$\mu_\dstb \gets \dstb^{(i)}$}
                $J^* \gets J^{(i)}$
            }
        }
        $\hat{V}(t_j,\state_j) \gets J^*$
    }
}
\Return{$D_{MPC} \gets D_{MPC} \cup (\state_j, t_j, \hat{\VV}(\state_j, t_j))$}
\PrintSemicolon
\end{algorithm}

\subsection{Sampling-Based MPC Dataset}
To generate the MPC dataset, we solve the following discrete-time version of the zero-sum game:
\begin{equation}
    \label{eq:MPCgame}
    \begin{aligned}
    \hat{\VV}(\state, \tinit)&=\min_{\dsig}\max _{\csig}  \min_{h \in\left\{0,1, \ldots, H+1\right\}} \ell\left(\state_h\right) \\
\text { s.t. } 
\state_{h+1} &=\bar{\f}(\state_h) + \bar{\fu}(\state_h, \ctrl_h) + \bar{\fd} (\state_h\,\dstb_h),\\
& \state_0=\state, \quad \ctrl_h \in \cfset, \quad \dstb_h \in \dfset.
\end{aligned}
\vspace{-0.5em}
\end{equation}

Here $h \in\{0,1, \ldots, H+1\}$ denotes the time steps between $\tinit$ and $\thor$ with $\traj:=\left[x_0, \ldots, x_{H+1}\right]$, $\csig:=\left[\ctrl_0, \ldots, \ctrl_H\right]$, and $\dsig:=\left[\dstb_0, \ldots, \dstb_H\right]$ the discrete state trajectory, control sequence and the disturbance sequence. The functions $\bar{\f}, \bar{\fu}, \bar{\fd}$ are the discretized dynamics that can be obtained from the continuous dynamics using first-order Euler approximation. 
This game can be solved in many ways; we will use sampling-based MPC (SB-MPC). Directly solving the resulting two-player game using MPC is challenging due to the need for co-optimization over both control and disturbance sequences. Instead, we generate two distinct datasets—one with SB-MPC controlling the agent while the disturbance follows a policy, and one with the roles reversed, allowing efficient learning of the value function. In practice, we generate two complementary datasets: 1) rollouts of SB-MPC optimized over control under a disturbance policy, and 2) rollouts of SB-MPC optimized over disturbance under a control policy. 
Our MPC algorithm is summarized in Alg. \ref{alg:robust_mpc_no_warmstart} and described below.

\subsubsection{Control Dataset} We first review how to generate the MPC dataset for the control input, shown in black and \textcolor{blue}{blue} text. The algorithm takes as inputs the MPC dataset size, $|D_{MPC}|$, time horizon $H$, and step size $\Delta t$, dynamics model $f$, current learned value function $\VV_\theta$, constraint function $\ell$, number of trajectories $N$, and number of refinement iterations $K$. We first uniformly sample states $x_0 \sim \mathcal{X}$ and times from the \ICRAmod{}{fixed} time horizon $\tint$.
Over the course of $K$ refinement iterations, we then generate $N$ trajectories starting at the sampled initial state $\state_0$. 
For each timestep $\Delta t$ over a horizon $H$, we first generate the control input by sampling a Gaussian distribution around a nominal mean control $\mu _{\ctrl, k}$, which varies per refinement iteration, and variance $\sigma_{u}^2$. In practice, we initialize it as $\mu _{\ctrl,0}=0$ and the variance is set to the maximum control bound. 
The robot then samples the disturbance by taking the action that minimizes the gradient of the current learned value function. 
The state, control, and disturbance are then propagated over $\Delta t$ via the dynamics.

Next, the objective function is set as the minimum of the constraint function over the trajectory. If the horizon $H$ is less than the total time horizon, the robot takes the min between the objective and the value function evaluated at the end of the horizon $H$ (i.e., cost-to-go). 
Finally, we update the best solution by making the new nominal control the best control action that we sampled so far. 
After $K$ refinement steps, the value function estimate at the initial state $V(x_0)$ is set to the highest cost of $N$ rollouts.

\subsubsection{Disturbance Dataset}
To generate the MPC dataset for the disturbance, we follow a similar procedure, shown in black and \textcolor{red}{red} text. 
The disturbance is now generated by sampling from a Gaussian distribution around a nominal mean $\mu _{\dstb,k}$ and variance $\sigma_\dstb^2$. 
The robot then samples the control by taking the action that maximizes the gradient of the current learned value function. After $K$ refinement steps, the initial state's value function's $V(x_0)$ estimate is set to the lowest cost of $N$ rollouts. 

These two datasets provide supervised learning signals for DeepReach with the following loss for both:
\begin{equation}
\begin{aligned}
    \mathcal{L}_\text{MPC} = \sum_{j=1}^{\lvert D_\text{MPC}\rvert} l_\text{MPC}(x_j,t_j,\hat{V}(x_j,t_j);\theta) \\
    l_\text{MPC}(x,t,\hat{V}(x,t);\theta) =\lVert \hat{V}(x,t) -V_\theta (x,t)\rVert_2,
    \end{aligned}
\end{equation}
with $\mathcal{L}_\text{MPC}=\mathcal{L}_\text{MPC,\textcolor{blue}{u}} + \mathcal{L}_\text{MPC,\textcolor{red}{d}}$ combining the sampling-based rollouts from the control and disturbance perspective.

\textit{Remark 1:}  
In addition to sampling-based MPC, standard Model Predictive Path Integral control (MPPI) could also be used to generate these datasets by sampling stochastic rollouts. Within our codebase, the user can select either SB-MPC or regular MPPI as the guiding policy; however, in practice, we use SB-MPC, which assigns full weight to the best rollout to approximate optimal control under uncertainty.

\subsection{Learning Value Function using MPC-Guided DeepReach}

To guide the learning of the value function over the full time horizon, we directly use lines 5-16 of Algorithm~2 in \cite{feng2025bridgingmodelpredictivecontrol}. 
This approach leverages a dataset of MPC rollouts together with PDE-based supervision to train 
a value function network that is amenable to disturbances. 
We briefly review key components of~\cite{feng2025bridgingmodelpredictivecontrol}, highlighting key differences. 

Firstly, as the MPC dataset collection, Alg.~\ref{alg:robust_mpc_no_warmstart}, relies on the gradient of the estimated value function, we do not consider a pretraining phase and collect the first MPC dataset after a set period of the self-supervised curriculum training. 

Next, Algorithm~2 of~\cite{feng2025bridgingmodelpredictivecontrol} relies on collecting data iteratively over a horizon that terminates at the current time in the curriculum $t_\text{curr}$ (hence starts from a time further in the curriculum). We employ the same technique but fix the evaluation of the learned value function's gradient for the policy generation to be evaluated at time $t_\text{curr}$ to remain in distribution with respect to the curriculum training. 

\subsection{A special pursuit-evasion filter for suboptimality in long-horizon games}
For two-player games with equally equipped agents, the value function is positive (safe for the evader) for the majority of the state space. Specifically, Player II's objective is to minimize the minimum cost over a pre-specified horizon, assuming an optimal Player I. 
As such, as the time-horizon increases, and the error of a learned system compounds, the approximated solution quality tends to deteriorate and erode Player II's performance. 

Therefore, we propose endowing Player II with a second policy as a filter, which prioritizes staying close to Player I when unable to ``catch'' Player I and switches to the classic pursuit-evasion policy when a ``catch'' is achievable. The ``following'' game has the following cost function:
\begin{equation}
\label{eq:objective_follow}
    J_\text{follow}(\state,\tinit, \csig, \dsig) = 
\max_{\tvar\in[\tinit, \thor]} \, \ell \big(\traj_{x,t}^{\csig,\dsig}(\tvar)\big),
\end{equation}
and the associated value function is:
\begin{equation}
\label{eq:valuefunc_follow}
    \VV_\text{follow}(\state,\tinit) = 
    \min_{\strategy[\csig] \in \Strategy}
    \max_{\csig \in \cset} J_\text{follow}(\state,\tinit, \csig, \dsig).
\end{equation}

This formulation is akin to~\cite{herbert2017fastrack}, which considers a worst-case tracking bound between two systems. While the resulting policy for the evader is not necessarily performant (e.g., it might be optimal under this policy for Player I to reach a state $x$ with $\ell(x)<0$ during the trajectory if this trajectory achieves a higher cost $J$ over the full trajectory), the policy for Player II tries to minimize the maximum boundary function, i.e. distance, over a trajectory. 
Such a policy induces Player II to stay close to Player I, but does not prioritize ``catching'' Player I. 
As such, we introduce a filtered strategy for Player II:
\begin{equation}
\dstb(x)=\begin{cases}
\arg\min\limits_\dstb H_d \> \> \> \> \>\text{if $\lvert \min\limits_\dstb H_d\rvert \geq \epsilon $}\\
\arg\min\limits_\dstb H_{d,\text{follow}},
\end{cases}
\end{equation}
where $H_d=\nabla \VV_\theta(x,T) \cdot w(x)d$ and $\epsilon \ll 1$. We denote this policy as \mad-FOLLOW for Player II.  

\section{Simulation Results}\label{sec:simulation}

\subsection{Simulation Performance Evaluation Metrics}\label{sec:metrics}

We benchmark our proposed approach against a set of baselines and evaluate performance across all case studies. The primary baselines and evaluation aspects include:

\textbf{Baselines:} Vanilla DeepReach \cite{bansal2021deepreach}, DP Grid-Based Methods \cite{FisacChenEtAl2015}, ISAACS \cite{hsu2023isaacs}, sampling-based MPC \cite{Sacks_2022}, and our Adversarial MPC-Guided Policies (\mad).  

\textbf{Metrics:} Recovered volume and IOU with ground truth for safe/unsafe regions and policy matchup tables: time-to-capture and capture rate under adversarial disturbances for evader/pursuer policy pairs.  

These metrics enable a comprehensive assessment of the strength of \mad, demonstrated by highly comparable performance to the ground truth solution (for low-dimensional settings) and outcompeting existing baselines' learned policies across all experiments.  

\subsection{Implementation Details}

For all baselines, we employ a three-layer sinusoidal neural network, a standard architecture in machine learning for capturing complex, high-frequency functions, with 512 neurons per layer. The networks are trained using the Adam optimizer with a learning rate of $\alpha = 2 \times 10^{-5}$ on an NVIDIA GeForce RTX 4090 GPU. The sampling-based MPC method uses a time step of $\Delta = 0.02s$ and 10 iterative sampling steps, with $N=100$ perturbed control sequences generated per step, with a dataset refinement horizon of $H_R=0.2$s for all experiments.

In each case study, we use the same number of training
iterations across all baselines to ensure a fair comparison.
However, Vanilla DeepReach does not
involve a fine-tuning phase, see~\cite{feng2025bridgingmodelpredictivecontrol}, as we observe that fine-tuning
encourages overfitting in these baselines and degrades their
performance. Detailed training parameters are provided in
Table \ref{tab:simMetrics}.

\begin{table}[h!]
\vspace{-.4cm}
\caption{Training time (compared with Vanilla DR), epochs, and MPC dataset size for \mad across systems.}
\centering
\resizebox{1.0\columnwidth}{!}{%
\begin{tabular}{lcccccc}
\hline
\textbf{System} & \textbf{\mad (hrs)} & \textbf{Vanilla (hrs)} & \textbf{Epochs} & \textbf{MPC Dataset Size}\\
\hline
Quadrotor (13D) & 4.2 & 1.7 & 110,000 & 10,000 \\
Dubins (6D)     & 3.6 & 1.5 & 150,000 & 20,000 \\
Drones (20D)    & 5.4 & 3.5 & 250,000 & 30,000 \\
\hline
\end{tabular}%
}
\label{tab:simMetrics}
\end{table}

\begin{figure}[t!]
    \centering
    \includegraphics[width=1.0\columnwidth]{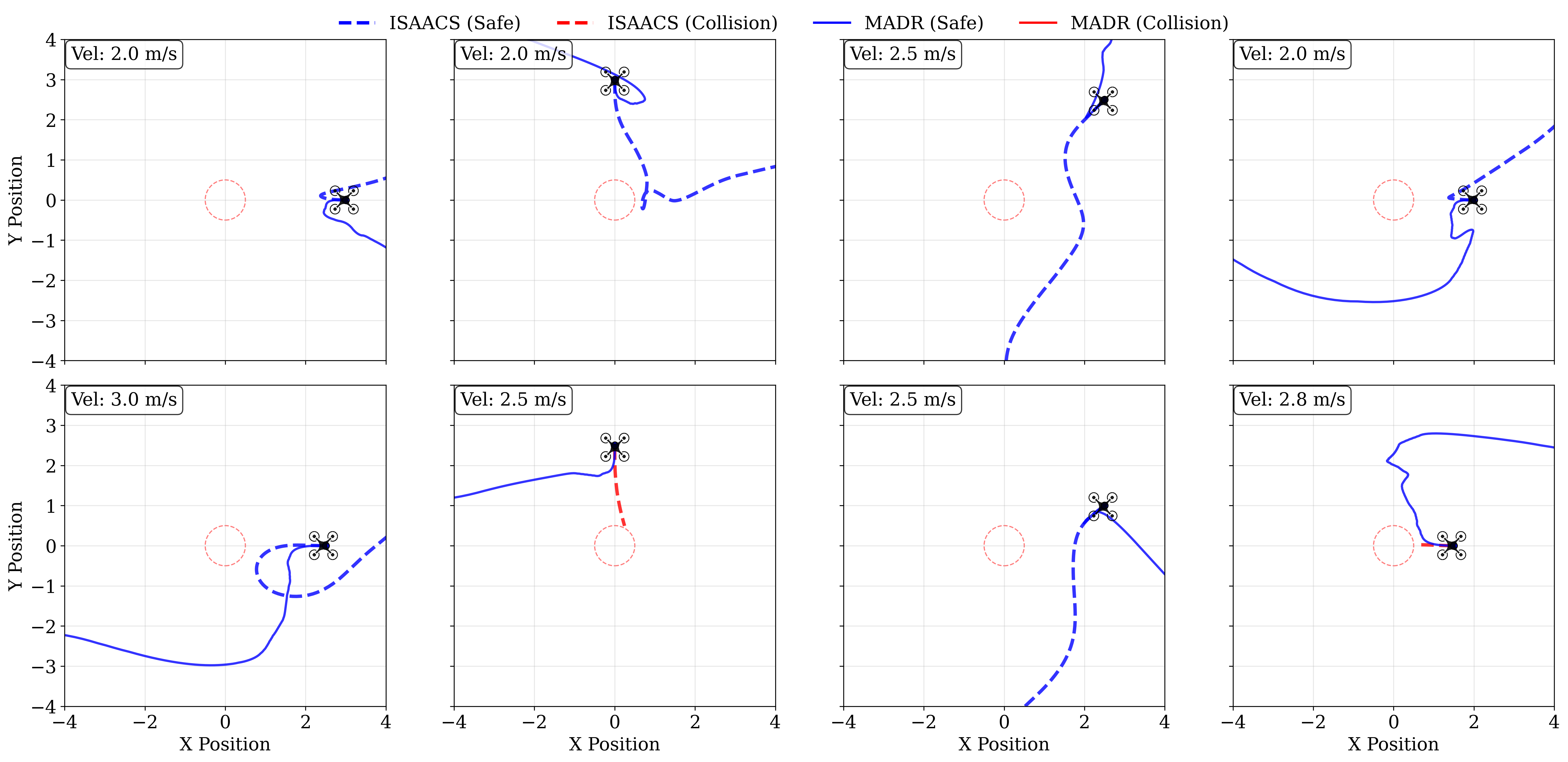} 
    \caption{Representative trajectory rollouts comparing \mad and ISAACS in the 13D quadrotor environment under wind disturbances. The solid lines correspond to \mad, while the dashed lines correspond to ISAACS.}
    \label{fig:avoid_cylinder_traj}
    \vspace{-1em}
\end{figure}

\subsection{13D Quadrotor with Disturbances}
We first consider a robustness scenario, simulating a full 13D, quaternion-based quadrotor under wind disturbances, including forces along the $x$, $y$, and $z$ axes and torques about the rotational axes. The drone must avoid a central pillar of radius 0.50 in the state space despite a high initial velocity. 

\begin{figure}[t!]
    \centering
    \includegraphics[width=0.8\columnwidth]{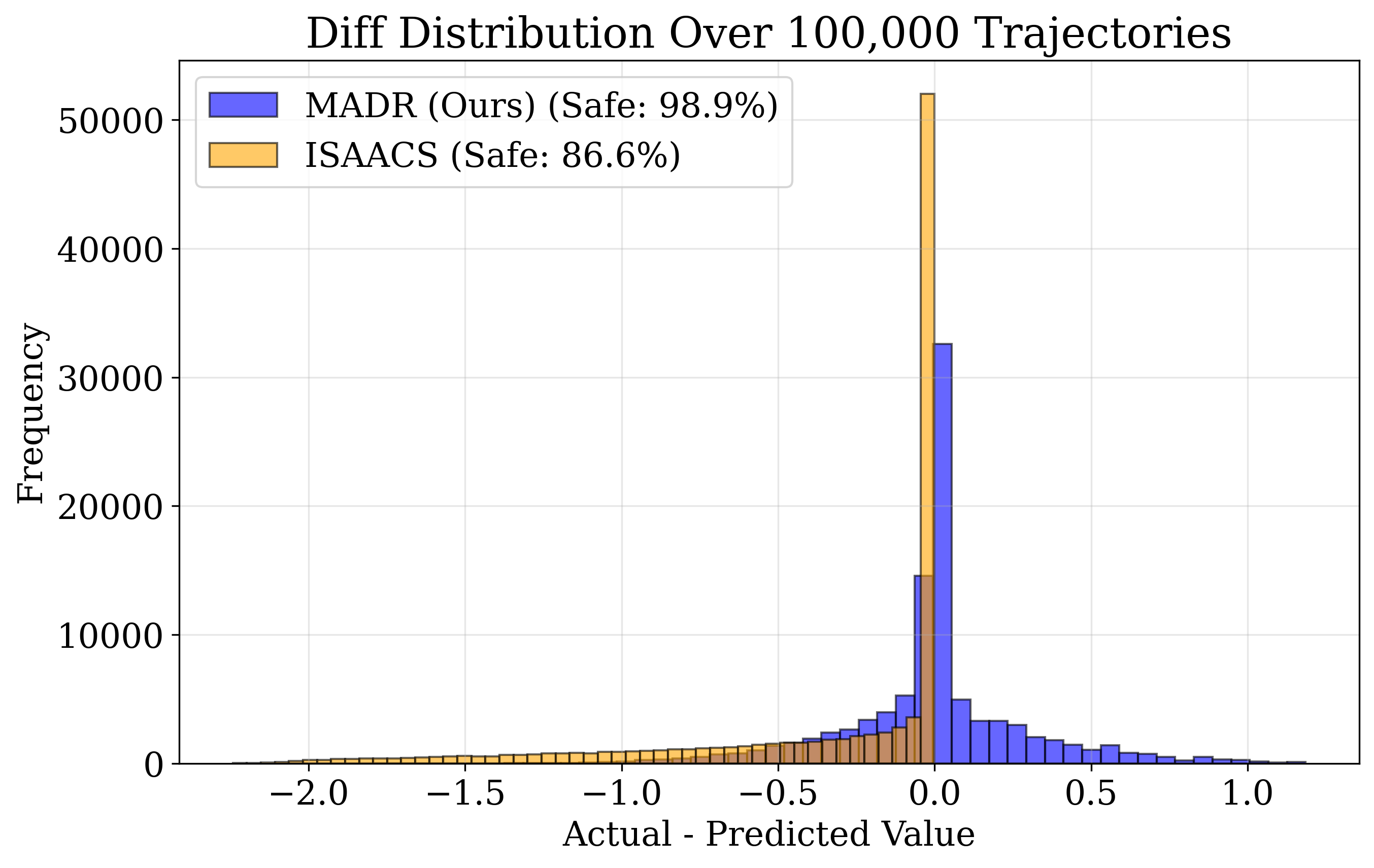} 
  \caption{Distribution of actual–predicted cost differences and safe rates over 100,000 trajectories, comparing \mad and ISAACS based on their  value functions. Negative values indicate overestimation of safety.}
    \vspace{-1em}
    \label{fig:cost_diff_quadrotor}
\end{figure}

Figure~\ref{fig:avoid_cylinder_traj} shows representative trajectory rollouts as the drone approaches the cylinder at high initial speeds, illustrating the effectiveness of \mad. Even under worst-case disturbances, our method consistently avoids the central pillar, whereas ISAACS fails in some cases. Figure~\ref{fig:cost_diff_quadrotor} presents the distribution of actual versus predicted trajectory costs for ISAACS and \mad, highlighting each method’s safety estimation. \mad achieves a 98.9\% safe rate compared with 86.6\% for ISAACS and provides significantly more accurate value function estimates, rarely overestimating safety. We believe this improvement comes from \mad being trained on higher-quality MPC data that better covers worst-case scenarios, yielding more reliable value estimates in safety-critical regions.


\subsection{6D Dubins Pursuit Evasion Game}
\label{sec:sim_dubins}

We then consider a zero-sum game for which we can obtain a ground truth comparison using dynamic programming (DP). Namely, we consider a Dubins pursuit–evasion game in which each player is described by three states: $x$, $y$, and $\theta$. The evader aims to escape the pursuer, which has a capture radius of 0.36, based on the TurtleBot's size. The players are constrained to a bounded state space of $[-3,3]$ in $x$ and $[-2,2]$ in $y$. Both players have a maximum turn rate of 1.9 rad/s and move at a constant velocity of 0.5 m/s. All methods are trained for a time horizon of 3 seconds, \ICRAmod{}{corresponding to the approximate convergence time of the Dubins reachability solution for the DP ground truth}.  

\begin{figure*}[h!]
    \centering
    \begin{minipage}[b]{\columnwidth}
        \centering
        \includegraphics[width=\columnwidth]{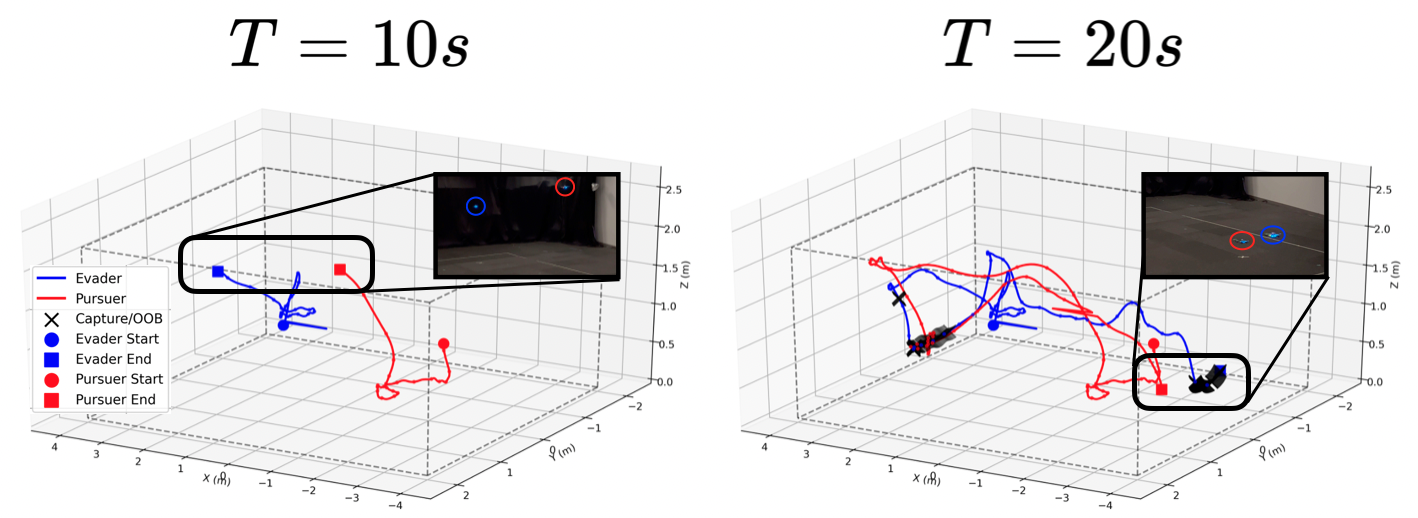}
    \end{minipage}
    \hfill
    \begin{minipage}[b]{\columnwidth}
        \centering
        \includegraphics[width=\columnwidth]{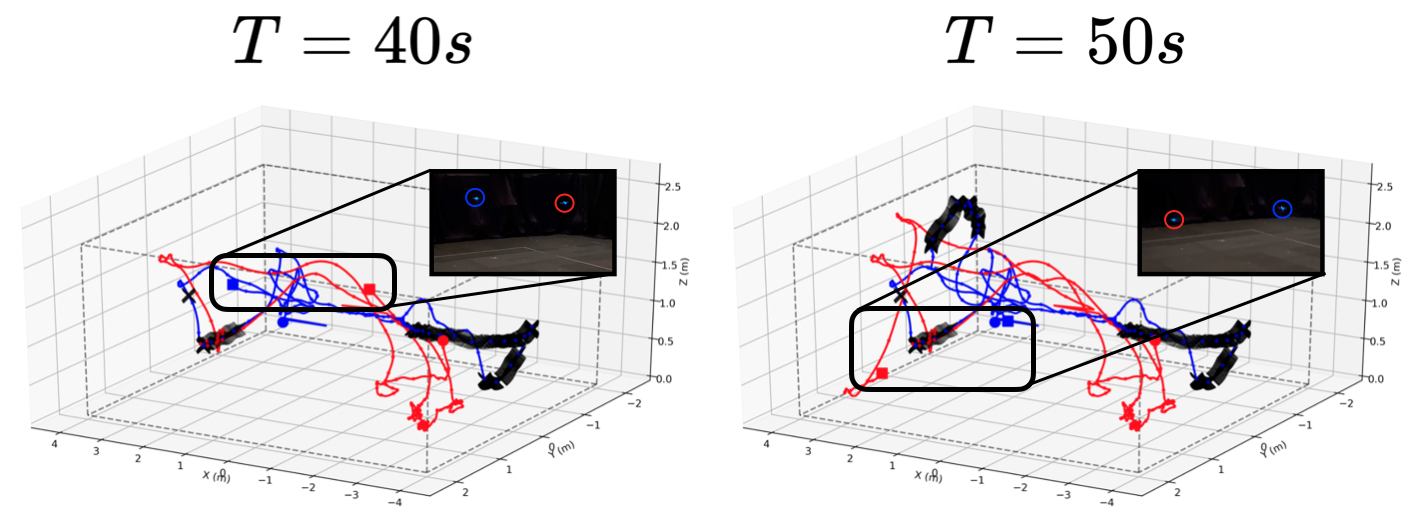}
    \end{minipage}

    \caption{A hardware demonstration of the drone vs. drone pursuit-evasion game with \mad. From left to right, four trajectory snapshots show the position  of the evader (blue) and the pursuer (red) up to that time point, with captures or out-of-bounds (OOB) marked by an 'x' (black). Both agents employ our policy, with the pursuer using the follow-filtered augmentation.}
    \label{fig:drone_v_drone_fig}
\vspace{-0.5cm}
\end{figure*}
\begin{table}[b]
\vspace{-0.4cm}
\caption{Comparison of methods for 6D Dubins: unsafe set volume within $x \in [-1.5, 1.5], y \in [-1.5, 1.5]$ and Intersection-over-Union (IOU) with DP ground truth. Higher IOU indicates better recovery of the true backward reachable tube (BRT).}
\centering
\resizebox{\columnwidth}{!}{%
\begin{tabular}{l|c|c}
\hline
\textbf{Method}   & \textbf{Unsafe Set Volume (\%)} & \textbf{IOU with Ground Truth (DP)} \\
\hline
DP (True BRT)     & 7.51  & 1.000  \\
\mad (Ours)       & 7.82  & 0.997 \\
Vanilla           & 4.60  & 0.969 \\
ISAACS            & 0.37  & 0.928 \\
\hline
\end{tabular}%
}
\vspace{-1em}
\label{tab:unsafe_iou}
\end{table}
Capture rates for each evader–pursuer policy pairing are summarized in Table~\ref{tab:safe_init_dubins}. As shown in the table, \mad performs near DP-optimal, significantly outperforming Vanilla DeepReach and ISAACS. Additionally, Table~\ref{tab:unsafe_iou} compares each method’s unsafe set volume and Intersection-over-Union (IOU) with the ground-truth value function, demonstrating that \mad comes within 0.03\% of the ground-truth solution. It maintains a slightly conservative unsafe set while outperforming ISAACS and Vanilla DeepReach in both accuracy and safe estimation.

\begin{figure}[t]
    \centering
    \includegraphics[width=1.0\columnwidth]{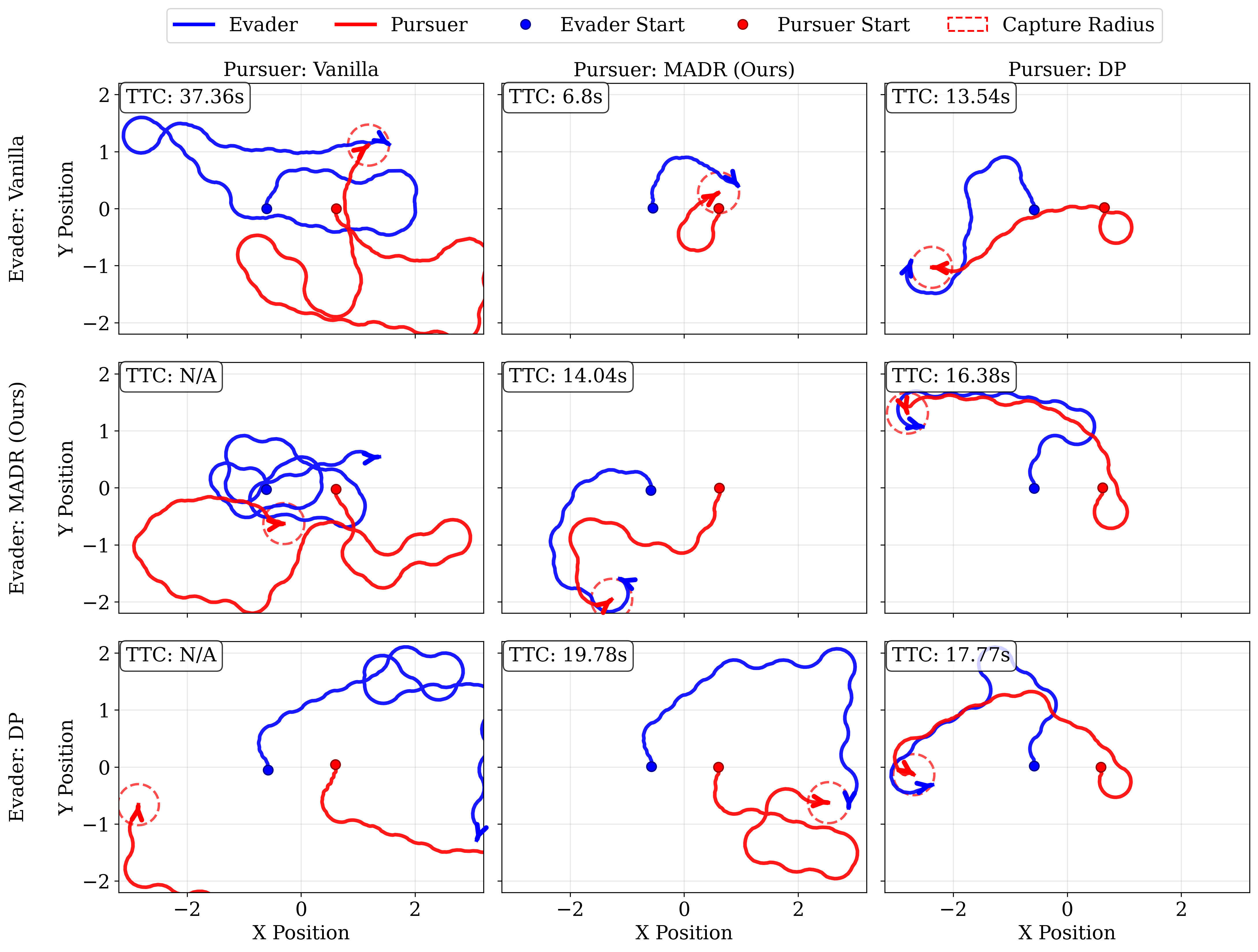} 
    \caption{TurtleBot hardware trajectories from one initial condition, showing qualitative differences across policies. Non-capture trajectories are truncated.}
    \label{fig:dubins_hardware_traj}
\vspace{-1.5em}
\end{figure}

\begin{table}[h!]
\caption{Capture rates (\%) over 100 safe initial states ($0 < V < 0.1$) based on the DP BRT ground truth for 6D Dubins. Rows correspond to Player I’s policy and columns to Player II’s policy. Cells are color-coded to highlight performance: blue indicates a stronger Evader, while red indicates a stronger Pursuer.}
\centering
\resizebox{\columnwidth}{!}{%
\begin{tabular}{c|c c c c c}
\hline
\diagbox{Evader ($\downarrow$)}{Pursuer ($\uparrow$)} & DP & \mad-FOLLOW & \mad & Vanilla & Isaacs \\
\hline
DP          & \colorcell{10} & \colorcell{6}  & \colorcell{10} & \colorcell{0}  & \colorcell{6}  \\
\mad & \colorcell{17} & \colorcell{13} & \colorcell{17} & \colorcell{1}  & \colorcell{7}  \\
Vanilla     & \colorcell{71} & \colorcell{70} & \colorcell{58} & \colorcell{45} & \colorcell{63} \\
Isaacs      & \colorcell{60} & \colorcell{56} & \colorcell{55} & \colorcell{20} & \colorcell{32} \\
\hline
\end{tabular}%
}
\label{tab:safe_init_dubins}
\vspace{-10pt}
\end{table}

\subsection{20D Drone Pursuit–Evasion Game}
\vspace{-4pt}
\label{sec:sim_drone}

In this scenario, we consider a drone pursuit–evasion game, where each drone is described by 10 states as detailed in~\cite{TonkensShinde2025EtAl}. The evader's capture boundary is defined by a halfellipse centered under the pursuer to represent the evader trying to avoid the downwash from the pursuer, similar to~\cite{li2025certifiable}. 
All policies are trained over a 1-second time horizon. 
The players are constrained to a bounded state space of $x\in[-4,4]$, $y\in[-2,2]$, and $z\in[0.2,2.0]$. Each drone has identical dynamics, with a maximum desired pitch/roll angle of 0.15~rad, maximum thrust of 14~N, and maximum linear and angular velocities of 2.0~m/s.

Table~\ref{tab:drone_capture} reports capture rates for each policy combination over 100 rollouts from random initial states. As illustrated, our evader successfully avoids capture from all policies except our own pursuer, while capturing all other policies. Moreover, the evader policy consistently outperforms all competing policies, and our pursuer performs comparably to MPC.

\begin{table}[b]
\caption{Capture rates (\%) in simulation for each evader–pursuer policy pairing in the 20D pursuit–evasion drone game, averaged over 100 rollouts from random initial conditions with a 1-second horizon.}
\centering 
\resizebox{\columnwidth}{!}{%
\begin{tabular}{c|c c c c}
\hline
\diagbox{Evader ($\downarrow$)}{Pursuer ($\uparrow$)} & \mad-FOLLOW & MPC & Vanilla & \mad \\
\hline
\mad & \colorlow{7} & \colorlow{1} & \colorlow{0} & \colorlow{4} \\
MPC         & \colorlow{25} & \colorlow{29} & \colorlow{21} & \colorlow{21}\\
Vanilla     & \colorlow{9} & \colorlow{11} & \colorlow{8} & \colorlow{10}\\
\hline
\end{tabular}%
}
\label{tab:drone_capture}
\end{table}

\section{Hardware Results}\label{sec:hardware}

Having validated our approach in simulation (Section~\ref{sec:simulation}), we now evaluate its performance in hardware to assess real-world scalability. 

\subsection{6D Dubins Pursuit Evasion Game}
To evaluate our approach in hardware, we implemented the 6D Dubins pursuit–evasion game on a pair of TurtleBots. Unlike the simulation studies (see Section~\ref{sec:sim_dubins}), which focused on short horizons of 3 seconds, the hardware experiments probe longer-term behavior and robustness. In particular, we consider rollouts lasting 500 seconds, allowing us to assess whether value functions learned over short horizons remain effective for longer-horizon interactions, capturing stability effects and strategic patterns not visible in brief simulations.

\begin{table}
\caption{Average time-to-capture (seconds) for each evader–pursuer policy pairing, evaluated over six selected initial states with large relative separation.}
\centering
\label{tab:dubins_hardware}
\resizebox{\columnwidth}{!}{%
\begin{tabular}{c|c c c c}
\hline
\diagbox{Evader ($\uparrow$)}{Pursuer ($\downarrow)$} & DP & \mad & \mad-FOLLOW & Vanilla \\
\hline
DP          & \cellcolor{lightred!70} 31 &  \cellcolor{lightblue!70} 84 & \cellcolor{lightred!60} 37 & \cellcolor{lightblue!90} 355 \\
\mad        & \cellcolor{lightred!90} 15 & \cellcolor{lightblue!30} 54 & \cellcolor{lightred!80} 25 & \cellcolor{lightblue!80} 243 \\
Vanilla     & \cellcolor{lightred!95} 11 & \cellcolor{lightred!95} 11 & \cellcolor{lightred!100} 8 & \cellcolor{lightblue!50} 72 \\
\hline
\end{tabular}%
}

\end{table}

Quantitative results on time to capture are reported in Table~\ref{tab:dubins_hardware}, with representative trajectories highlighting qualitative differences in strategy shown in Figure~\ref{fig:dubins_hardware_traj}. As shown, our policy significantly outperforms Vanilla DeepReach and achieves performance comparable to DP optimal, with similar times to capture. The trajectories further demonstrate long-term planning on the pursuer's side, successfully cornering evaders. Despite the short-horizon learned value function, our method produces meaningful results even over 500-second experiments.

\subsection{20D Drone Hardware Platform}

To ground the 20D pursuit–evasion formulation in a realistic setting, we implement our framework on a pair of Crazyflie 2.1 quadrotors, each modeled with the 10-dimensional state space (See Section~\ref{sec:sim_drone}). 

All neural network queries are performed offboard on a workstation. The Crazyflies receive only low-level combined thrust and desired attitude commands via a wireless link at 50~Hz. 
For precise localization during experiments, the drones operate in a motion-capture arena providing real-time position and velocity measurements, which is fused with the internal IMU to provide full-state observability. 
Each Crazyflie uses its onboard PID-based attitude controller to track the commanded thrust and torques. The actuation limits are consistent with the Crazyflie hardware: maximum thrust of 15~N per rotor, roll and pitch angles constrained to $\pm 0.15$~rad, and angular velocity limits of 2.0~rad/s. 

We plot a representative roll-out, showcasing long-horizon behavior in Figure~\ref{fig:drone_v_drone_fig}. We compared the standard pursuit-evasion \mad policy to \mad-FOLLOW, finding that across near and distant initial positions \mad-FOLLOW achieved capture for 26.4\% of the 3-minute trajectory on average versus 11.6\% for \mad. Qualitatively, both policies induce diving attacks and dodge evasions (pictured in Figure~\ref{fig:frontpage}), but the follow-filtered pursuit policy manages to approach the evader more quickly when far away. See the website for the trajectory videos.

\subsection{Humanoid–Drone Pursuit–Evasion Experiments}

To interrogate the robustness of our method, we conducted an experiment facing \mad against a Unitree G1 humanoid driven by a human-operator. In this game, we model the humanoid with quadrotor-like dynamics, although the opponent is weaker in reality due to its dynamic limitations, but one that we consider conservatively. A Crazyflie mounted on the humanoid provided state estimation via motion capture, while the aerial drone agent executed the \mad-FOLLOW strategy for the pursuer role and \mad for the evader role. 

\begin{figure}[!ht]
    \centering
    \includegraphics[width=1.\linewidth]{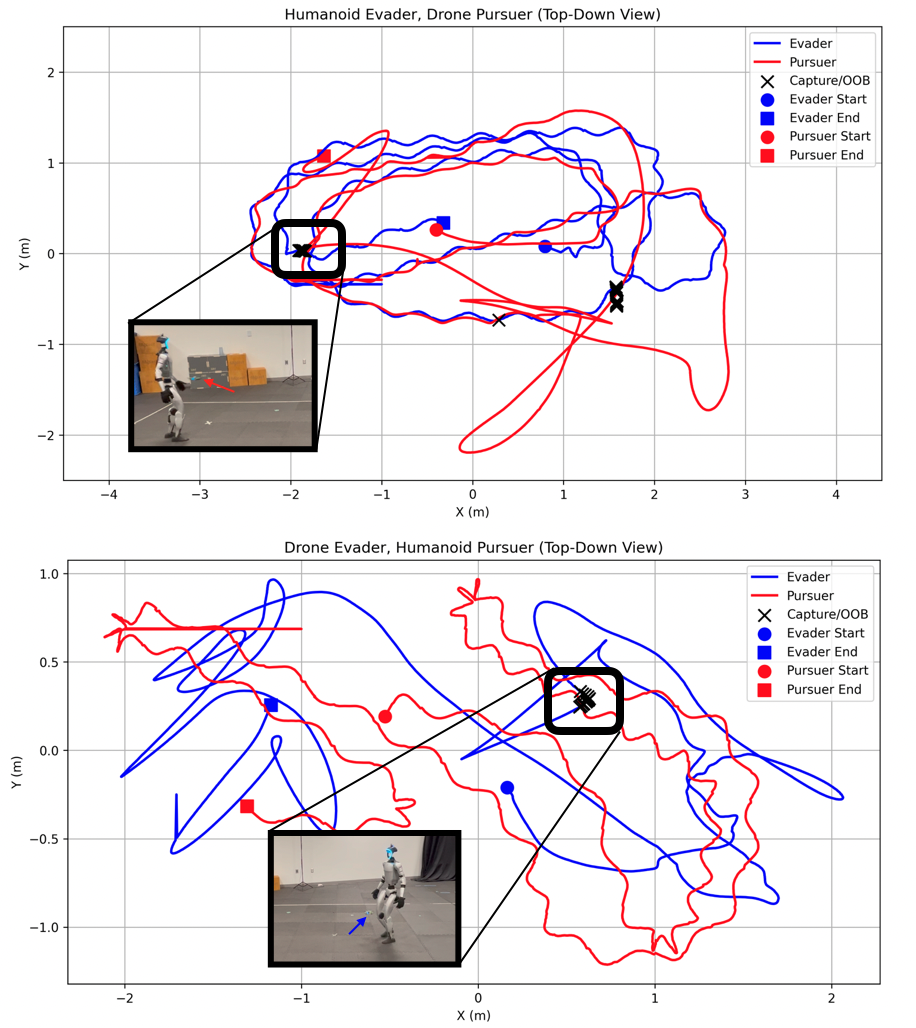}
    \caption{Two hardware demonstrations of the teleoperated humanoid vs. \mad drone pursuit-evasion game. On top, the teleoperated humanoid evades, while on the bottom, it pursues.
    The positions of the evader (blue) and the pursuer (red) are plotted with captures marked as `x' (black). }
    \label{fig:humanoid}
    \vspace{-1em}
\end{figure}

Two of six example trajectories are plotted in Figure~\ref{fig:humanoid}, and the full videos are on the website. In the pursuit case, the drone autonomously follows the rear of the humanoid and at another point rapidly lunges at its abdomen (Figure~\ref{fig:humanoid}, top). In the evader case, the drone frequents the lower heights, tending to ``side-step'' the approaching humanoid, but is occasionally knee'd (Figure~\ref{fig:humanoid}, bottom).

\section{Conclusions}\label{sec:conclusion}

In this work, we proposed \mad, a robust learning framework that augments the DeepReach process with supervision from adversarial model predictive control (MPC). By leveraging worst-case control–disturbance interactions as expert guidance, \mad enables accurate approximation of Hamilton–Jacobi (HJ) reachability solutions and backward reachable tubes (BRTs) in the presence of both disturbances and adversarial agents.  
Our results highlight two key contributions. First, we show that \mad scales effectively across a range of dynamical systems and dimensions, and validate its practicality through both simulation and hardware experiments. Second, we demonstrate that \mad robustly handles diverse zero-sum games, from worst-case environmental disturbances to pursuit-evasion games. 
Together, these results underscore \mad's potential for real-world deployment in safety-critical autonomous systems, where reliable decision-making under uncertainty is essential.   
\ICRAmod{}{Future work will focus on establishing a stronger theoretical grounding for \mad, including analysis of convergence, approximation error bounds, and formal safety guarantees that preserve the properties of the underlying HJ reachability formulation. }

\bibliographystyle{IEEEtran}
\bibliography{main, ASL_papers, SASLab}

@Preamble{"\newcommand{\noopsort}[1]{} " #
"\newcommand{\printfirst}[2]{#1} " #
"\newcommand{\singleletter}[1]{#1} " #
"\newcommand{\switchargs}[2]{#2#1} "}

@String{jrn_AR_CRAS               = {{Annual Review of Control, Robotics, and Autonomous Systems}}}

@String{jrn_IEEE_RAL                   = {{IEEE Robotics and Automation Letters}}}

@String { jrn_IEEE_TAC              = {{IEEE Transactions on Automatic Control}} }

@String { jrn_IFAC_A                = {{Automatica}} }

@String{jrn_IFAC_online                = {{IFAC-Papers Online}}}

@String { proc_ACM_HSCC             = {{Hybrid Systems: Computation and Control}} }

@String{proc_CoRL                      = {{Conf.\ on Robot Learning}}}

@String { proc_ICML                 = {{Int.\ Conf.\ on Machine Learning}} }

@String { proc_IEEE_ACC             = {{American Control Conference}} }

@String { proc_IEEE_CDC             = {{Proc.\ IEEE Conf.\ on Decision and Control}} }

@String { proc_IEEE_ICRA            = {{Proc.\ IEEE Conf.\ on Robotics and Automation}} }

@String{proc_L4DC                      = {{Learning for Dynamics \& Control Conference}}}

@String{proc_NIPS                      = {{Conf.\ on Neural Information Processing Systems}}}

@String { proc_RSS                  = {{Robotics: Science and Systems}} }

@String { proc_WAFR                 = {{Workshop on Algorithmic Foundations of Robotics}} }

@String { pub_CUP                   = {{Cambridge Univ.\ Press}} }

@String{jrn_IEEE_RAL              = {{IEEE Robotics and Automation Letters}}}

@String{proc_L4DC                      = {{Learning for Dynamics \& Control}}}

@Inproceedings{FisacLugovoyEtAl2019,
  author    = {Fisac, J. F. and Lugovoy, N. F. and {Rubies-Royo}, V. and Ghosh, S. and Tomlin, C. J.},
  title     = {Bridging Hamilton-Jacobi Safety Analysis and Reinforcement Learning},
  booktitle = proc_IEEE_ICRA,
  year      = {2019},
}

@article{BrunkeGreefEtAl2021,
  title={Safe learning in robotics: From learning-based control to safe reinforcement learning},
  author={Brunke, Lukas and Greeff, Melissa and Hall, Adam W and Yuan, Zhaocong and Zhou, Siqi and Panerati, Jacopo and Schoellig, Angela P},
  journal=jrn_AR_CRAS,
  volume={5},
  number={1},
  pages={411--444},
  year={2022},
}

@article{papavassilopoulos1979nash,
  title={On the existence of Nash strategies and solutions to coupled Riccati equations in linear-quadratic games},
  author={Papavassilopoulos, George P. and Medanic, J.V. and Cruz, J.B.},
  journal={Journal of Optimization Theory and Applications},
  volume={28},
  number={1},
  pages={49--76},
  year={1979},
  doi={10.1007/BF00933600},
  publisher={Springer}
}

@article{Zhao_2023,
   title={Stackelberg Meta-Learning Based Control for Guided Cooperative LQG Systems},
   volume={56},
   number={2},
   journal=jrn_IFAC_online,
   author={Zhao, Yuhan and Zhu, Quanyan},
   year={2023},
   pages={10120–10125} }

@article{falconi2025distributionally,
  title={Distributionally robust LQG control under distributed uncertainty},
  author={Falconi, Lucia and Ferrante, Augusto and Zorzi, Mattia},
  journal=jrn_IFAC_A,
  volume={174},
  pages={112128},
  year={2025},
  publisher={Elsevier}
}

@INPROCEEDINGS{robustMPC,
  author={Campo, Peter J. and Morari, Manfred},
  booktitle=proc_IEEE_ACC, 
  title={Robust Model Predictive Control}, 
  year={1987},
}

@inproceedings{Sacks_2022,
   title={Learning to Optimize in Model Predictive Control},
   booktitle=proc_IEEE_ICRA, 
   author={Sacks, Jacob and Boots, Byron},
   year={2022}, }

@inproceedings{TonkensShinde2025EtAl,
    author = {Sander Tonkens and Nikhil U. Shinde and Azra Begzadic and Michael C. Yip and Jorge Cortes and Sylvia L. Herbert},
    title = {From Space to Time: Enabling Adaptive Safety with Learned Value Functions via Disturbance Recasting},
    booktitle = proc_CoRL,
    year = {2025}
}

@article{li2025certifiable,
  title={Certifiable Reachability Learning Using a New Lipschitz Continuous Value Function},
  author={Li, Jingqi and Lee, Donggun and Lee, Jaewon and Dong, Kris Shengjun and Sojoudi, Somayeh and Tomlin, Claire},
  journal=jrn_IEEE_RAL,
  year={2025},
  volume={10},
  number={4},
  pages={3582-3589},
}

@inproceedings{williams2016mppi,
  title={Model predictive path integral control: From theory to parallelized implementation},
  author={Williams, Grady and Aldrich, Andrew and Theodorou, Evangelos A},
  booktitle=proc_IEEE_ICRA,
  year={2016},
}

@inproceedings{lowe2017multi,
  title={Multi-agent actor-critic for mixed cooperative-competitive environments},
  author={Lowe, Ryan and Wu, Yi and Tamar, Aviv and Harb, Jean and Abbeel, Pieter and Mordatch, Igor},
  booktitle=proc_NIPS,
  year={2017}
}

@Book{BorrelliBemporadEtAl2017,
  author    = {Borrelli, F. and Bemporad, A. and Morari, M.},
  title     = {Predictive Control for Linear and Hybrid Systems},
  publisher = pub_CUP,
  year      = {2017},
}

@Inproceedings{FisacChenEtAl2015,
  author    = {Fisac, J. F. and Chen, M. and Tomlin, C. J. and Sastry, S. S.},
  title     = {Reach-avoid problems with time-varying dynamics, targets and constraints},
  booktitle = proc_ACM_HSCC,
  year      = {2015},
}

@Inproceedings{BansalChenEtAl2017b,
  author    = {Bansal, S. and Chen, M. and Herbert, S. and Tomlin, C. J.},
  title     = {{Hamilton-Jacobi} reachability: A brief overview and recent advances},
  booktitle = proc_IEEE_CDC,
  year      = {2017},
}

@inproceedings{wang2025magicsadversarialrlminimax,
      title={MAGICS: Adversarial RL with Minimax Actors Guided by Implicit Critic Stackelberg for Convergent Neural Synthesis of Robot Safety}, 
      author={Justin Wang and Haimin Hu and Duy Phuong Nguyen and Jaime Fernández Fisac},
      year={2024},
      booktitle=proc_WAFR
}

@inproceedings{Borquez_2023,
   title={Parameter-Conditioned Reachable Sets for Updating Safety Assurances Online},
   booktitle=proc_IEEE_ICRA,
   author={Borquez, Javier and Nakamura, Kensuke and Bansal, Somil},
   year={2023}, }

@inproceedings{nguyen2024gameplayfiltersrobustzeroshot,
      title={Gameplay Filters: Robust Zero-Shot Safety through Adversarial Imagination}, 
      author={Duy P. Nguyen and Kai-Chieh Hsu and Wenhao Yu and Jie Tan and Jaime F. Fisac},
      year={2024},
      booktitle=proc_CoRL
}

@inproceedings{
	so2024solving,
	title={Solving Minimum-Cost Reach Avoid using Reinforcement Learning},
	author={So, Oswin and Ge, Cheng and Fan, Chuchu},
	booktitle=proc_NIPS,
	year={2024},
}

@article{schwenzer2021review,
  title={Review on model predictive control: An engineering perspective},
  author={Schwenzer, Max and Ay, Muzaffer and Bergs, Thomas and Abel, Dirk},
  journal={The international journal of advanced manufacturing technology},
  volume={117},
  number={5},
  pages={1327--1349},
  year={2021},
  publisher={Springer}
}

@article{mitchell2005HJ,
  author    = {I. M. Mitchell and A. M. Bayen and C. J. Tomlin},
  title     = {A time-dependent Hamilton-Jacobi formulation of reachable sets for continuous dynamic games},
  journal   = jrn_IEEE_TAC,
  volume    = {50},
  number    = {7},
  pages     = {947--957},
  year      = {2005}
}

@inproceedings{chen2016fast,
  author    = {M. Chen and S. Herbert and C. J. Tomlin},
  title     = {Fast reachable set approximations via state decoupling disturbances},
  booktitle = proc_IEEE_CDC,
  year      = {2016}
}

@inproceedings{fisac2015reachavoid,
  author    = {J. F. Fisac and A. Akametalu and M. Zeilinger and S. Kaynama and J. Gillula and C. Tomlin},
  title     = {Reach-avoid problems with time-varying dynamics, targets and constraints},
  booktitle = proc_ACM_HSCC,
  year      = {2015}
}

@inproceedings{herbert2017fastrack,
  author    = {S. Herbert and M. Chen and S. Han and W. Ma and C. J. Tomlin},
  title     = {FaSTrack: A modular framework for real-time motion planning with guaranteed tracking},
  booktitle = proc_IEEE_CDC,
  year      = {2017}
}

@inproceedings{bansal2021deepreach,
  author    = {S. Bansal and C. J. Tomlin},
  title     = {DeepReach: A Deep Learning Approach to High-Dimensional Reachability},
  booktitle = proc_IEEE_ICRA,
  year      = {2021}
}

@inproceedings{pinto2017robust,
  author    = {Lerrel Pinto and James Davidson and Rahul Sukthankar and Abhinav Gupta},
  title     = {Robust Adversarial Reinforcement Learning},
  booktitle = proc_ICML,
  year      = {2017},
}

@INPROCEEDINGS{li2023robust,
  author={Li, Zeyang and Hu, Chuxiong and Li, Shengbo Eben and Cheng, Jia and Wang, Yunan},
  booktitle=proc_IEEE_CDC, 
  title={Robust Safe Reinforcement Learning under Adversarial Disturbances}, 
  year={2023},
}

@inproceedings{hsu2023isaacs,
  author    = {Judy Hsu and Jason Nguyen and Jaime F. Fisac},
  title     = {ISAACS: Iterative Soft Adversarial Actor-Critic for Safety},
  booktitle = proc_L4DC,
  year      = {2023},
}

@inproceedings{feng2025bridgingmodelpredictivecontrol, title={Bridging Model Predictive Control and Deep Learning for Scalable Reachability Analysis}, author={Zeyuan Feng and Le Qiu and Somil Bansal}, year={2025}, booktitle=proc_RSS }

@inproceedings{sharpless2024linear,
  title={Linear Supervision for Nonlinear, High-Dimensional Neural Control and Differential Games},
  author={Sharpless, William and Feng, Zeyuan and Bansal, Somil and Herbert, Sylvia},
  booktitle=proc_L4DC,
  year={2025}
}

@inproceedings{fridovich2020efficient,
  title={Efficient iterative linear-quadratic approximations for nonlinear multi-player general-sum differential games},
  author={Fridovich-Keil, David and Ratner, Ellis and Peters, Lasse and Dragan, Anca D and Tomlin, Claire J},
  booktitle=proc_IEEE_ICRA,
  year={2020},
}

\end{document}